\documentclass[11pt,a4paper]{article}
\usepackage[hyperref]{acl2020}
\usepackage{times}
\usepackage{latexsym}

\usepackage{microtype}

\aclfinalcopy 
\def\aclpaperid{3136} 

\usepackage{amsmath}
\DeclareMathOperator{\sign}{sign}

\usepackage{caption}
\makeatletter

\usepackage{setspace}

\usepackage{xspace}
\newcommand\BLEU{\textsc{Bleu}\xspace}
\newcommand\TER{\textsc{Ter}\xspace}
\usepackage{graphicx}
\usepackage{wrapfig}

\title{Successfully Applying the Stabilized Lottery Ticket Hypothesis to the Transformer Architecture}

\author{Christopher Brix, Parnia Bahar, Hermann Ney\\
  Human Language Technology and Pattern Recognition Group
Computer Science Department \\
RWTH Aachen University \\
D-52056 Aachen, Germany \\
\texttt{<surname>@i6.informatik.rwth-aachen.de}}

\date{}

\begin{document}
\maketitle
\begin{abstract}
Sparse models require less memory for storage and enable a faster inference by reducing the necessary number of FLOPs.
This is relevant both for time-critical and on-device computations using neural networks.
The stabilized lottery ticket hypothesis states that networks can be pruned after none or few training iterations, using a mask computed based on the unpruned converged model. 
On the transformer architecture and the WMT 2014 English$\to$German and English$\to$French tasks, we show that stabilized lottery ticket pruning performs similar to magnitude pruning for sparsity levels of up to 85\%, and propose a new combination of pruning techniques that outperforms all other techniques for even higher levels of sparsity. 
Furthermore, we confirm that the parameter's initial sign and not its specific value is the primary factor for successful \mbox{training}, and show that magnitude pruning \mbox{could} be used to find winning lottery tickets.
\end{abstract}

\section{Introduction}
Current neural networks are heavily growing in depth, with many fully connected layers.
As every fully connected layer includes large matrices, models often contain millions of parameters.
This is commonly seen as an over-parameterization \citep{Dauphin-overparameterization, Denil-overparam-predictable-weights}.
Different techniques have been proposed to decide which weights can be pruned.
In structured pruning techniques (\citealp{voita-pruning-heads}), whole neurons or even complete layers are removed from the network.
Unstructured pruning only removes individual connections between neurons of succeeding layers, keeping the global network architecture intact.
The first technique directly results in smaller model sizes and faster inference, while the second offers more flexibility in the selection of which parameters to prune.
Although the reduction in necessary storage space can be realized using sparse matrix representations \citep{stanimirovic-csc}, most popular frameworks currently do not have sufficient support for sparse operations.
However, there is active development for possible solutions \citep{Liu-sparse-framework2, han-sparse-framework, elsen-fast-convnet}.
This paper compares and improves several unstructured pruning techniques.
The main contributions of this paper are to:
\begin{itemize}
    \item verify that the stabilized lottery ticket hypothesis \citep{frankle-stabilized-lt} performs similar to magnitude pruning \cite{narang-mp-iteratively} on the transformer architecture \citep{vaswani-attention-is-all-you-need} with 60M parameters up to a sparsity of 85\%, while magnitude pruning is superior for higher sparsity levels.
    \item demonstrate significant improvements for high sparsity levels over magnitude pruning by using it in combination with the lottery ticket hypothesis.
    \item confirm that the signs of the initial parameters are more important than the specific values to which they are reset, even for large networks like the transformer.
    \item show that magnitude pruning could be used to find winning lottery tickets, i.e., the final mask reached using magnitude pruning may be an indicator for which initial weights are most important.
\end{itemize}

\section{Related Work}
\citet{han-magnitude-pruning} propose the idea of pruning weights with a low magnitude to remove connections that have little impact on the trained model.
\citet{narang-mp-iteratively} incorporate the pruning into the main training phase by slowly pruning parameters during the training, instead of performing one big pruning step at the end.
\citet{zhu-prune-or-not-prune} provide an implementation for magnitude pruning in networks designed using the tensor2tensor software \citep{tensor2tensor}.

\citet{frankle-lottery-ticket} propose the lottery ticket hypothesis, which states that dense networks contain sparse sub-networks that can be trained to perform as good as the original dense model.
They find such sparse sub-networks in small architectures and simple image recognition tasks and show that these sub-networks might train faster and even outperform the original network.
For larger models, \citet{frankle-stabilized-lt} propose to search for the sparse sub-network not directly after the initialization phase, but after only a few training iterations.
Using this adapted setup, they are able to successfully prune networks having up to 20M parameters.
They also relax the requirement for lottery tickets so that they only have to beat randomly initialized models with the same sparsity level.

\citet{uber-signs} show that the signs of the weights in the initial model are more important than their specific values.
Once the least important weights are pruned, they set all remaining parameters to fixed values, while keeping their original sign intact. 
They show that as long as the original sign remains the same, the sparse model can still train more successfully than one with a random sign assignment.
\citet{frankle-early-phase} reach contradicting results for larger architectures, showing that random initialization with original signs hurts the performance.

\citet{graves-state-of-sparsity} compare different pruning techniques on challenging image recognition and machine translation tasks and show that magnitude pruning achieves the best sparsity-accuracy trade-off while being easy to implement.

In concurrent work, \citet{Yu-playing} test the stabilized lottery ticket on the transformer architecture and the WMT 2014 English$\to$German task, as well as other architectures and fields.

This paper extends the related works by demonstrating and comparing the applicability of different pruning techniques on a deep architecture for two translation tasks, as well as proposing a new combination of pruning techniques for improved performance.

\section{Pruning Techniques}
\label{sec:pruningTechniques}
In this section, we give a brief formal definition of each pruning technique.
For a more detailed description, refer to the respective original papers.

In the given formulas, a network is assumed to be specified by its parameters $\theta$.
When training the network for $T$ iterations, $\theta_t$ for $t \in [0,T]$ represents the parameters at timestep $t$.

\paragraph{Magnitude Pruning (MP)}
relies on the magnitude of parameters to decide which weights can be pruned from the network.
Different techniques to select which parameters are selected for pruning have been proposed \citep{collins-mp-variant1, han-magnitude-pruning, guo-mp-variant2, zhu-prune-or-not-prune}.
In this work, we rely on the implementation from \citet{zhu-prune-or-not-prune} where the parameters of each layer are sorted by magnitude, and during training, an increasing percentage of the weights are pruned.
It is important to highlight that MP is the only pruning technique not requiring multiple training runs.

\paragraph{Lottery Ticket (LT)}
pruning assumes that for a given mask $m$, the initial network $\theta_0$ already contains a sparse sub-network $\theta_0 \odot m$ that can be trained to the same accuracy as $\theta_0$.
To determine $m$, the parameters of each layer in the converged model $\theta_T$ are sorted by magnitude, and $m$ is chosen to mask the smallest ones such that the target sparsity $s_T$ is reached.
We highlight that even though $m$ is determined using $\theta_T$, it is then applied to $\theta_0$ before the sparse network is trained.
To reach high sparsity without a big loss on accuracy, \citet{frankle-lottery-ticket} recommend to prune iteratively, by training and resetting multiple times.

\paragraph{Stabilized Lottery Ticket (SLT)}
pruning is an adaptation of LT pruning for larger models.
\citet{frankle-stabilized-lt} propose to apply the computed mask $m$ not to the initial model $
\theta_0$, but to an intermediate checkpoint $\theta_t$ where $0 < t \ll T$ is chosen to be early during the training. 
They recommend to use $0.001T \leq t \leq 0.07T$ and refer to it as iterative magnitude pruning with rewinding.
We highlight that \citet{frankle-stabilized-lt} always choose $\theta_t$ from the first, dense model, while this work choses $\theta_t$ from the last pruning iteration.

\paragraph{Constant Lottery Ticket (CLT)}
pruning assumes that the specific random initialization is not important.
Instead, only the corresponding choice of signs affects successful training.
To show this, \citet{uber-signs} propose to compute $\theta_t \odot m$ as in SLT pruning, but then to train $f(\theta_t \odot m)$ as the sparse model.
Here, $f$ sets all remaining parameters $p$ in each layer $l$ to $\sign(p) \cdot \alpha_l$, i.e., all parameters in each layer have the same absolute value, but their original sign.
In all of our experiments, $\alpha_l$ is chosen to be $\alpha_l = \sqrt{\frac{6}{n_{l_{in}} + n_{l_{out}}}}$ where $n_{l_{in}}$ and $n_{l_{out}}$ are the respective incoming and outgoing connections to other layers.

\paragraph{SLT-MP}
is a new pruning technique, proposed in this work.
It combines both SLT pruning and MP in the following way:
First, SLT pruning is used to find a mask $m$ with intermediate sparsity $s_i$.
This might be done iteratively.
$\theta_t \odot m$ with sparsity $s_i$ is then used as the initial model for MP (i.e., $\theta'_0 = \theta_t \odot m$).
Here, in the formula for MP, $s_0 = s_i$.
We argue that this combination is beneficial, because in the first phase, SLT pruning removes the most unneeded parameters, and in the second phase, MP can then slowly adapt the model to a higher sparsity.

\paragraph{MP-SLT}
is analogue to SLT-MP:
First, MP is applied to compute a trained sparse network $\theta_T$ with sparsity $s_i$.
This trained network directly provides the corresponding mask $m$.
$\theta_t \odot m$ is then used for SLT pruning until the target sparsity is reached.
This pruning technique tests whether MP can be used to find winning lottery tickets.

\section{Experiments}
\label{sec:experiments}
We train the models on the WMT 2014 English$\to$German and English$\to$French datasets, consisting of about 4.5M and 36M sentence pairs, respectively.
\texttt{newstest2013} and \texttt{2014} are chosen to be the development and test sets.

All experiments have been performed using the base transformer architecture as described in \citep{vaswani-attention-is-all-you-need}.\footnote{Using the hyperparameters in \texttt{transformer\_base\_v3} in https://github.com/tensorflow/tensor2tensor/\\blob/838f1a99e24a9391a8faf6603e90d476444110a0/\\tensor2tensor/models/transformer.py with the corresponding adaptations for TPUs.}
The models are trained for 500k iterations on a single v3-8 TPU, saving checkpoints every 25k iterations.
For all experiments, we select the best model based on the \BLEU{} score on the development set.
For MP, we only evaluate the last 4 checkpoints, as earlier checkpoints do not have the targeted sparsity.
Intermediate MP sparsity levels $s_t$ are computed as $s_t = s_T + \min\{0, (s_0 - s_T) \large(1 - \frac{t}{400000})^3\}$ \citep{zhu-prune-or-not-prune}.
For efficiency reasons, weights are only pruned every 10k iterations.
Unless stated otherwise, we start with initial sparsity $s_0 = 0$.
The final sparsity \mbox{$s_T$ is individually given for each experiment.}

We prune only the matrices, not biases.
We report the approximate memory consumption of all trained models using the Compressed Sparse Column (CSC) format \citep{stanimirovic-csc}, which is the default for sparse data storage in the SciPy toolkit \citep{scipy}.

Our initial experiments have shown that Adafactor leads to an improvement of 0.5 \BLEU{} compared to Adam.
Hence, we select it as our optimizer with a learning rate of $lr(t) = \frac{1}{\max(t, w)}$ for $w = 10$k warmup steps.
We note that this differs from the implementation by \citet{graves-state-of-sparsity}, in which Adam has been used.
We highlight that for all experiments that require a reset of parameter values (i.e., LT, SLT, CLT, SLT-MP, and MP-SLT), we reset $t$ to $0$, to include the warmup phase \mbox{in every training run.}

A shared vocabulary of 33k tokens based on word-pieces \citep{wu-wordpieces} is used.
The reported case-sensitive, tokenized \BLEU{} scores are computed using SacreBLEU \citep{post-sacrebleu}, \TER{} scores are computed using MultEval \citep{clark-multeval}.
All results are averaged over two separate training runs.
For all experiments that require models to be reset to an early point during training, we select a checkpoint after 25k iterations.

All iterative pruning techniques except SLT-MP are pruned in increments of 10 percentage points up to 80\%, then switching to 5 points increments, and finally pruning to 98\% sparsity.
SLT-MP is directly trained using SLT pruning to 50\% and further reduced by SLT to 60\%, \mbox{before switching to MP.}

\begin{table*}
\centering
\begin{tabular}{r@{\hskip3pt}r|c@{\hskip3pt}c@{\hskip7pt}c@{\hskip3pt}c@{\hskip7pt}c@{\hskip3pt}c@{\hskip7pt}c@{\hskip3pt}c@{\hskip7pt}c@{\hskip3pt}c@{\hskip7pt}c@{\hskip3pt}c}
\hline
\textbf{Sparsity} & \textbf{Memory} & \multicolumn{2}{c}{\textbf{MP}} & \multicolumn{2}{c}{\textbf{LT}} & \multicolumn{2}{c}{\textbf{SLT}} & \multicolumn{2}{c}{\textbf{CLT}} & \multicolumn{2}{c}{\textbf{SLT-MP}}& \multicolumn{2}{c}{\textbf{MP-SLT}} \\
 & & \BLEU & \TER & \BLEU & \TER & \BLEU & \TER & \BLEU & \TER & \BLEU & \TER & \BLEU & \TER \\
\hline
0\% & 234 MB & 26.8 & 64.5 & 26.8 & 64.5 & 26.8 & 64.5 & 26.8 & 64.5 & 26.8 & 64.5 & \textcolor{gray}{26.8} & \textcolor{gray}{64.5} \\
10\% & 226 MB & 26.8 & \textbf{64.5} & 26.7 & 64.6 & 26.8 & 64.9 & \textbf{26.9} & 64.7 & n/a & n/a & \textcolor{gray}{26.8} & \textcolor{gray}{64.5} \\
20\% & 206 MB & 26.7 & \textbf{64.5} & 26.2 & 65.3 & 26.9 & 64.6 & \textbf{27.0} & \textbf{64.5} & n/a & n/a & \textcolor{gray}{26.7} & \textcolor{gray}{64.5} \\
30\% & 184 MB & 26.4 & 65.0 & 26.0 & 65.3 & \textbf{26.9} & 64.8 & \textbf{26.9} & \textbf{64.7} & n/a & n/a & \textcolor{gray}{26.4} & \textcolor{gray}{65.0} \\
40\% & 161 MB & 26.5 & \textbf{64.8} & 25.8 & 65.7 & \textbf{27.1} & 65.1 & 26.8 & 65.0 & n/a & n/a & \textcolor{gray}{26.5} & \textcolor{gray}{64.8} \\
50\% & 137 MB & 26.4 & 65.0 & 25.4 & 66.3 & 26.6 & 65.2 & \textbf{26.7} & 65.2 & 26.4$^\dagger$ & \textbf{64.9}$^\dagger$ & \textcolor{gray}{26.4} & \textcolor{gray}{65.0} \\
60\% & 112 MB & 25.9 & 65.5 & 24.9 & 66.5 & 26.4 & 65.7 & \textbf{26.8} & \textbf{65.0} & 26.4$^\dagger$ & 65.1$^\dagger$ & \textcolor{gray}{25.9} & \textcolor{gray}{65.5} \\
70\% & 86 MB & 25.7 & 65.8 & 24.2 & 67.6 & 25.6 & 66.9 & \textbf{26.2} & 65.8 & \textbf{26.2}$^\ddag$ & \textbf{65.3}$^\ddag$ & 25.6 & 66.0 \\
80\% & 59 MB & 24.8 & 66.8 & 23.2 & 68.4 & 24.8 & 67.7 & 24.1 & 67.9 & \textbf{25.6}$^\ddag$ & \textbf{65.9}$^\ddag$ & 24.6 & 67.2 \\
85\% & 46 MB & 23.9 & 67.7 & 22.3 & 69.8 & 23.7 & 68.5 & 23.7 & 68.0 & \textbf{24.9}$^\ddag$ & \textbf{66.4}$^\ddag$ & 23.9 & 67.9 \\
90\% & 31 MB & 22.9 & 69.0 & 20.9 & 72.0 & 21.7 & 71.4 & 21.6 & 70.6 & \textbf{23.5}$^\ddag$ & \textbf{68.4}$^\ddag$ & 22.4 & 69.8 \\
95\% & 17 MB & 20.2 & 72.9 & 18.1 & 75.4 & 17.4 & 77.1 & 18.2 & 73.3 & \textbf{20.5}$^\ddag$ & \textbf{72.3}$^\ddag$ & 18.5 & 75.5 \\
98\% & 7 MB & 15.8 & 78.9 & 13.3 & 81.2 & 11.0 & 86.9 & 14.6 & \textbf{78.2} & \textbf{16.1}$^\ddag$ & 79.2$^\ddag$ & 13.5 & 82.6 \\
\hline
\end{tabular}
\caption{\label{results-de}
En$\to$De translation: \BLEU [\%] and \TER [\%] scores of the final model at different sparsity levels, evaluated on \texttt{newstest2014}.
For SLT-MP, models marked with $\dagger$ are trained with SLT pruning, models marked with $\ddag$ are trained with MP.
For MP-SLT, the MP model with 60\% sparsity was used for SLT pruning.
For each sparsity level, the best score is highlighted.
}
\end{table*}

\begin{table}[h!]
\centering
\begin{tabular}{@{\hskip3pt}r@{\hskip3pt}|@{\hskip3pt}c@{\hskip3pt}c@{\hskip4pt}c@{\hskip3pt}c@{\hskip4pt}c@{\hskip3pt}c@{\hskip4pt}c@{\hskip1pt}c@{}}
\hline
\textbf{Sp.} & \multicolumn{2}{c@{\hskip10pt}}{\textbf{MP}} & \multicolumn{2}{c@{\hskip10pt}}{\textbf{SLT}} & \multicolumn{2}{c@{\hskip10pt}}{\textbf{CLT}} & \multicolumn{2}{c}{\phantom{\,\,Slt-MP}\llap{\textbf{SLT-MP}}}\\
 & \small\BLEU & \small\TER & \small\BLEU & \small\TER & \small\BLEU & \small\TER & \small\BLEU & \small\TER \\
\hline
0\% & 39.3 & 57.2 & 39.3 & 57.2 & 39.3 & 57.2 & 39.3 & 57.2 \\
10\% & 39.3 & \textbf{57.2} & 39.3 & 57.4 & \textbf{39.4} & 57.4 & n/a & n/a \\
20\% & \textbf{39.3} & 57.2 & \textbf{39.3} & \textbf{57.1} & \textbf{39.3} & 57.2 & n/a & n/a \\
30\% & 39.3 & 57.1 & \textbf{39.8} & \textbf{56.7} & 39.7 & 56.9 & n/a & n/a \\
40\% & 38.8 & 57.8 & \textbf{39.7} & \textbf{56.9} & 39.2 & 57.3 & n/a & n/a \\
50\% & 38.8 & 57.7 & 39.2 & 57.4 & \textbf{39.4} & 57.4 & 39.0$^\dagger$ & \textbf{57.3}$^\dagger$ \\
60\% & 38.5 & 57.9 & 39.0 & 57.6 & \textbf{39.2} & 57.5 & \textbf{39.2}$^\dagger$ & \textbf{57.4}$^\dagger$ \\
70\% & 38.2 & 58.4 & 38.4 & 58.3 & \textbf{38.9} & \textbf{57.8} & 38.5$^\ddag$ & 58.2$^\ddag$ \\
80\% & 37.5 & 59.1 & 37.4 & 59.3 & 37.3 & 59.2 & \textbf{38.0}$^\ddag$ & \textbf{58.7}$^\ddag$ \\
85\% & 37.0 & \textbf{59.6} & 36.9 & \textbf{59.6} & 35.7 & 61.1 & \textbf{37.4}$^\ddag$ & \textbf{59.6}$^\ddag$ \\
90\% & 35.6 & 61.4 & 34.7 & 62.1 & 33.7 & 62.9 & \textbf{35.9}$^\ddag$ & \textbf{60.4}$^\ddag$ \\
95\% & 32.7 & 63.8 & 28.5 & 68.0 & 29.6 & 65.7 & \textbf{33.1}$^\ddag$ & \textbf{63.1}$^\ddag$ \\
98\% & 27.1 & 69.6 & \makebox[0pt]{\phantom{$^\ast$}21.8$^\ast$} & \makebox[0pt]{\phantom{$^\ast$}73.9$^\ast$} & 19.6 & 75.9 & \textbf{27.3}$^\ddag$ & \textbf{68.9}$^\ddag$ \\
\hline
\end{tabular}
\caption{\label{results-fr}
En$\to$Fr translation: \BLEU [\%] and \TER [\%] scores of the final model at different sparsity levels, evaluated on \texttt{newstest2014}.
For SLT-MP, models marked with $\dagger$ are trained with SLT pruning, models marked with $\ddag$ are trained with MP.
($\ast$) indicates a result of a single run, as the second experiment failed.
For each sparsity level, the best score is highlighted.
}
\end{table}

\section{Experimental Results}
In this section, we evaluate the experimental results for English$\to$German and English$\to$French translation given in Tables \ref{results-de} and \ref{results-fr} to provide a comparison between the different pruning techniques described in \mbox{Section \ref{sec:pruningTechniques}}.

\paragraph{MP}
Tables \ref{results-de} and \ref{results-fr} clearly show a trade-off between accuracy and network performance.
For every increase in sparsity, the performance degrades accordingly.
We especially note that even for a sparsity of 50\%, the baseline performance cannot be achieved.
In contrast to all other techniques in this paper, MP does not require any reset of parameter values.
Therefore, the training duration is not increased.

\paragraph{LT}
\citet{frankle-lottery-ticket} test the LT hypothesis on the small ResNet-50 architecture \citep{he-resnet50} which is applied to ImageNet \citep{russakovsky-imagenet}.
\citet{graves-state-of-sparsity} apply LT pruning to the larger transformer architecture and the translation task WMT 2014 English$\to$German, noting that it has been outperformed by MP.
As seen in Table \ref{results-de}, simple LT pruning is outperformed by MP at all sparsity levels.
Because LT pruning is an iterative process, training a network with sparsity 98\% requires to train and reset the model 13 times, causing a big training overhead without any gain in performance.
Therefore, simple LT pruning cannot be recommended for complex architectures.

\paragraph{SLT}
The authors of the SLT hypothesis \citep{frankle-stabilized-lt} state that after 0.1-7\% of the training, the intermediate model can be pruned to a sparsity of 50-99\% without serious impact on the accuracy.
As listed in Tables \ref{results-de} and \ref{results-fr}, this allows the network to be pruned up to 60\% sparsity without a significant drop in \BLEU, and is on par with MP up to 85\% sparsity.

As described in Section \ref{sec:experiments}, for resetting the models, a checkpoint after $t=25$k iterations is used.
For a total training duration of 500k iterations, this amounts to 5\% of the training and is therefore within the 0.1-7\% bracket given by \citet{frankle-stabilized-lt}.
For individual experiments, we have also tried $t \in \{12.5\text{k}, 37.5\text{k}, 500\text{k}\}$ and have gotten similar results to those listed in this paper.
It should be noted that for the case $t = 500$k, SLT pruning becomes a form of MP, as no reset happens anymore.
We propose a more thorough hyperparameter search for the optimal $t$ value \mbox{as future work}.

Importantly, we note that the magnitude of the parameters in both the initial and the final models increases with every pruning step.
This causes the model with 98\% sparsity to have weights greater than 100, making it unsuitable for checkpoint averaging, as the weights become too sensitive to minor changes.
\citet{Yu-playing} report that they do successfully apply checkpoint averaging.
This might be because they choose $\theta_t$ from the dense training run for resetting, while we choose $\theta_t$ from the most recent sparse training.

\paragraph{CLT}
The underlying idea of the LT hypothesis is, that the untrained network already contains a sparse sub-network which can be trained individually.
\citet{uber-signs} show that only the signs of the remaining parameters are important, not their specific random value.
While \citet{uber-signs} perform their experiments on MNIST and CIFAR-10, we test this hypothesis on the WMT 2014 English$\to$German translation task using a deep transformer architecture.

Surprisingly, CLT pruning outperforms SLT pruning on most sparsity levels (see Table \ref{results-de}).
By shuffling or re-initializing the remaining parameters, \citet{frankle-lottery-ticket} have already shown that LT pruning does not just learn a sparse topology, but that the actual parameter values are of importance.
As the good performance of the CLT experiments indicates that changing the parameter values is of little impact as long as the sign is kept the same, we verify that keeping the original signs is indeed necessary.
To this end, we randomly assign signs to the parameters after pruning to 50\% sparsity. After training, this model scores 24.6\% \BLEU{} and 67.5\% \TER{}, a clear performance degradation from the 26.7\% \BLEU{} and 65.2\% \TER{} given in Table \ref{results-de}.
Notably, this differs from the results by \citet{frankle-early-phase}, as their results indicate that the signs alone are not enough to guarantee good performance.

\paragraph{SLT-MP}
Across all sparsity levels, the combination of SLT pruning and MP outperforms all other pruning techniques.
For high sparsity values, \mbox{SLT-MP} models are also superior to the SLT models by \citet{Yu-playing}, even though they start of from a better performing baseline.
We hypothesize that by first discarding 60\% of all parameters using SLT pruning, MP is able to fine-tune the model more easily, because the least useful parameters are already removed.

We note that the high weight magnitude for sparse SLT models prevents successful MP training.
Therefore, we have to reduce the number of SLT pruning steps by directly pruning to 50\% in the first pruning iteration.
However, as seen by comparing the scores for 50\% and 60\% sparsity on SLT and SLT-MP, this does not hurt the SLT performance.

For future work, we suggest trying different sparsity values $s_i$ for the switch between SLT and MP.

\paragraph{MP-SLT}
Switching from MP to SLT pruning causes the models to perform slightly better than for pure SLT pruning.
This indicates that MP may be useful to find winning lottery tickets.

\section{Conclusion}
In conclusion, we have shown that the stabilized lottery ticket (SLT) hypothesis performs similar to magnitude pruning (MP) on the complex transformer architecture up to a sparsity of about 85\%.
Especially for very high sparsities of 90\% or more, MP has proven to perform reasonably well while being easy to implement and having no additional training overhead.
We also have successfully verified that even for the transformer architecture, only the signs of the parameters are important when applying the SLT pruning technique.
The specific initial parameter values do not significantly influence the training.
By combining both SLT pruning and MP, we can improve the sparsity-accuracy trade-off.
In SLT-MP, SLT pruning first discards 60\% of all parameters, so MP can focus on fine-tuning the model for maximum accuracy.
Finally, we show that MP could be used to determine winning lottery tickets.

In future work, we suggest performing a hyperparameter search over possible values for $t$ in SLT pruning (i.e., the number of training steps that are not discarded during model reset), and over $s_i$ for the switch from SLT to MP in SLT-MP.
We also recommend looking into why CLT pruning works in our setup, while \citet{frankle-early-phase} present opposing results.

\paragraph{Acknowledgements}
We would like to thank the anonymous reviewers for their valuable feedback.
\begin{wrapfigure}{l}{0.1\textwidth}
\vspace{-5mm}
\begin{center}
\includegraphics[width=0.12\textwidth]
{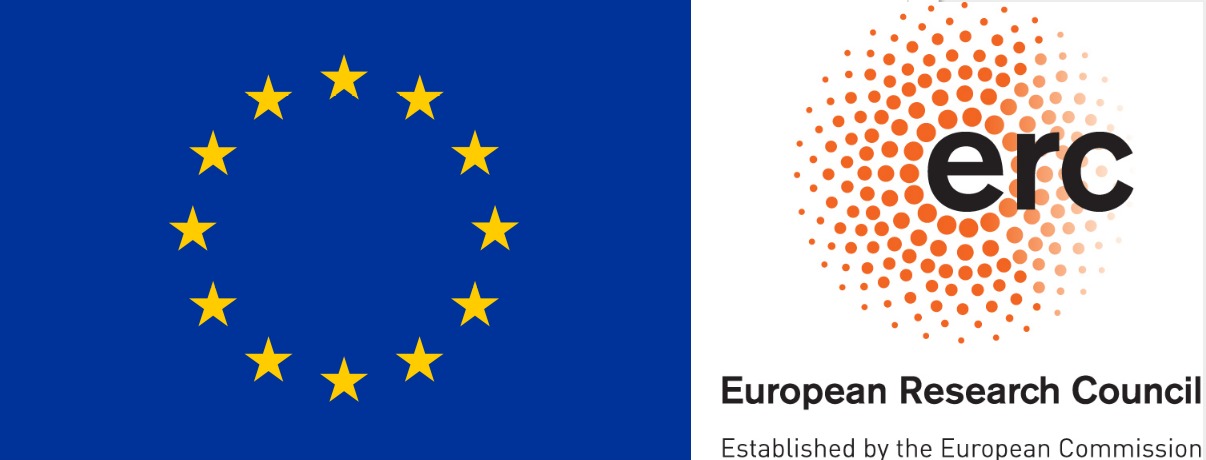} \\
\vspace{2mm}
\includegraphics[width=0.12\textwidth]
{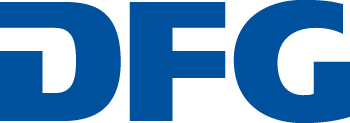}
\end{center}
\vspace{-4mm}
\end{wrapfigure}
This work has received funding from the European Research Council (ERC) under the European Union's Horizon 2020 research and innovation programme (grant agreement No 694537, project "SEQCLAS"), the Deutsche Forschungsgemeinschaft (DFG; grant agreement NE 572/8-1, project "CoreTec").
Research supported with Cloud TPUs from Google's TensorFlow Research Cloud (TFRC).
The work reflects only the authors' views and none of the funding parties is responsible for any use that may be made of the information it contains.

\bibliography{acl2020}
\bibliographystyle{acl_natbib}

\end{document}